%% file: main.tex
\documentclass[10pt,twocolumn,letterpaper]{article}
\usepackage{iccv}              


\definecolor{iccvblue}{rgb}{0.21,0.49,0.74}
\usepackage[pagebackref,breaklinks,colorlinks,allcolors=iccvblue]{hyperref}

\usepackage{times}
\usepackage{epsfig}
\usepackage{graphicx}
\usepackage{amsmath}
\usepackage{amssymb}
\usepackage{multirow}
\usepackage{booktabs}
\usepackage{float}
\usepackage{siunitx}
\usepackage{adjustbox}
\usepackage{enumitem}

\def\paperID{3068} 
\def\confName{ICCV}
\def\confYear{2025}

\title{\ AerialVG: A Challenging Benchmark for Aerial Visual Grounding by \\ Exploring Positional Relations}

\author{
  Junli Liu\thanks{ Authors contributed equally} $^{ 1,2}$ \quad Qizhi Chen$^{\ast2,3}$  \quad Zhigang Wang$^{\ast2}$ \quad Yiwen Tang$^{1,2}$ \quad Yiting Zhang$^{2}$ \\ Chi Yan$^2$ \quad
  Dong Wang$^2$  \quad Xuelong Li$^{4}$ \quad Bin Zhao \thanks{Corresponding author} $^{1,2}$\\ 
  \normalsize $^1$Northwestern Polytechnical University \quad $^2$Shanghai AI Laboratory \quad $^3$Zhejiang University \quad $^4$TeleAI  
}

\begin{document}

\makeatletter
\let\@oldmaketitle\@maketitle
\renewcommand{\@maketitle}{\@oldmaketitle
  \vspace{-6ex}
  \begin{center}
  \captionsetup{type=figure}
  \setcounter{figure}{0}
  \includegraphics[trim=0ex 0 0 0, clip, width=0.93\textwidth]{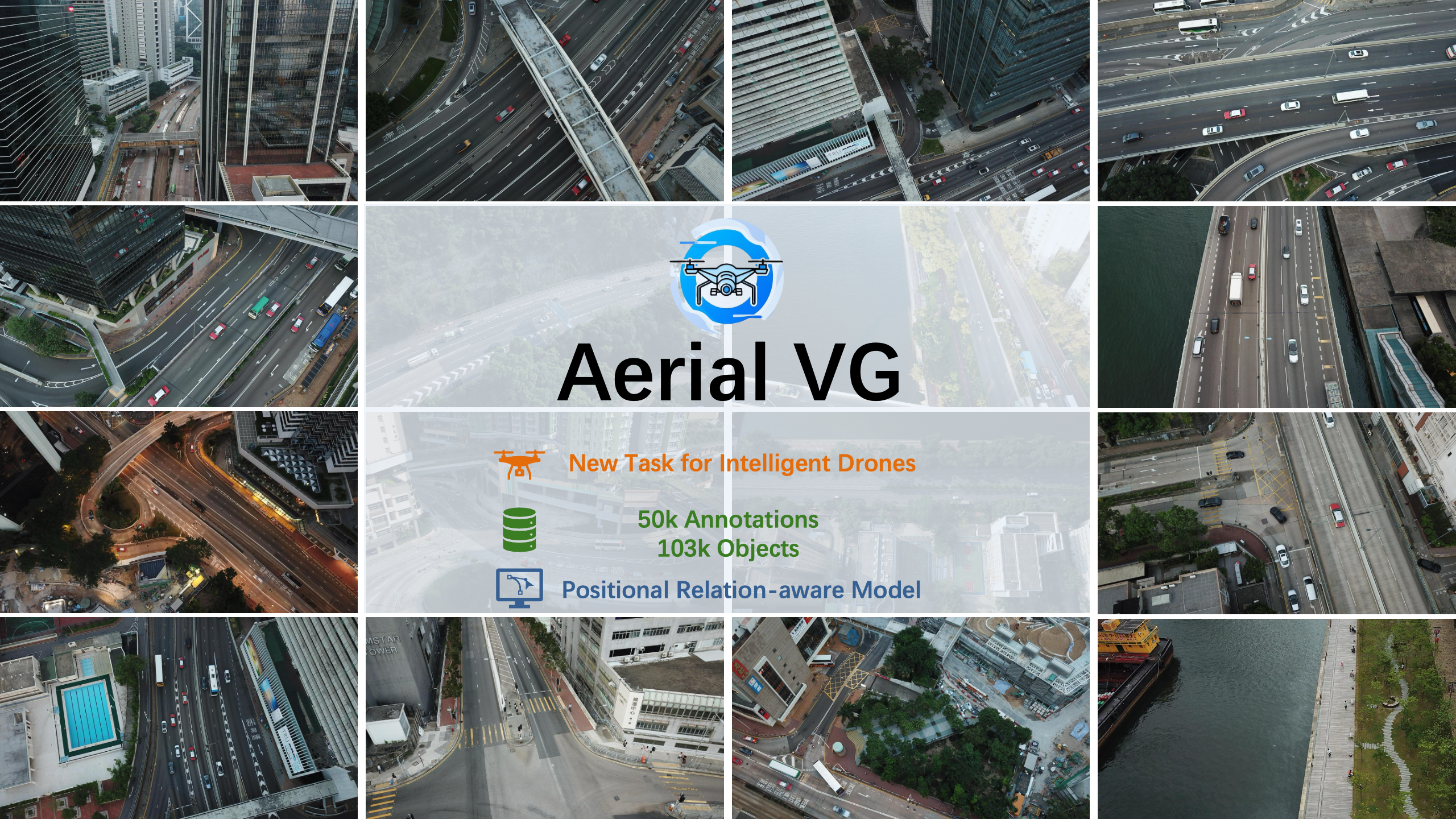}
    \caption{\textbf{Overview of AerialVG.} This work introduces aerial visual grounding (AerialVG), a new task for intelligent drones. We provide the first dataset for this task, consisting of 50K annotations and 103K objects. Additionally, an AerialVG model is proposed to explore the positional relations among objects and achieve superior context-aware visual grounding in drone applications.
    } 
    \label{fig:AerialVG}
  \end{center}
}

\makeatother

\maketitle
\input{sec/0_abstract}    
\input{sec/1_intro}

\input{sec/2_related_work}
\input{sec/3_dataset}

\input{sec/4_Method}

\input{sec/5_experiment}
\input{sec/6_conclusion}


{
    \small
    \bibliographystyle{ieeenat_fullname}
    \bibliography{main}
}


\end{document}

%% file: sec/0_abstract.tex
\begin{abstract}




Visual grounding (VG) aims to localize target objects in an image based on natural language descriptions. In this paper, we propose AerialVG, a new task focusing on visual grounding from aerial views. Compared to traditional VG, AerialVG poses new challenges, \emph{e.g.}, appearance-based grounding is insufficient to distinguish among multiple visually similar objects, and positional relations should be emphasized. Besides, existing VG models struggle when applied to aerial imagery, where high-resolution images cause significant difficulties. 
To address these challenges, we introduce the first AerialVG dataset, consisting of 5K real-world aerial images, 50K manually annotated descriptions, and 103K objects. Particularly, each annotation in AerialVG dataset contains multiple target objects annotated with relative spatial relations, requiring models to perform comprehensive spatial reasoning.
Furthermore, we propose an innovative model especially for the AerialVG task, where a Hierarchical Cross-Attention is devised to focus on target regions, and a Relation-Aware Grounding module is designed to infer positional relations. Experimental results validate the effectiveness of our dataset and method, highlighting the importance of spatial reasoning in aerial visual grounding. The code will be  released at \href{url}{https://github.com/Ideal-ljl/AerialVG}.

\end{abstract}

%% file: sec/1_intro.tex
\section{Introduction}

Visual Grounding (VG) is a challenging task that involves identifying specific regions or objects within an image following natural language instructions. This task requires models to deeply understand both visual and textual modalities, effectively aligning image context with corresponding linguistic expressions. VG is critical for a wide range of applications, including visual question answering, vision-and-language navigation, and human-robot interaction, \emph{etc} \cite{vgrs, zhu2020vision, Qu_fast, yan2023universal, NEURIPS2024_16ce2c09, Qu_2024_CVPR}.

With the continuous progress of multimodal models, the performance of VG tasks has been significantly improved. However, existing models and datasets still have obvious limitations. Current research focuses primarily on ground objects, and the exploration of aerial visual grounding remains underdeveloped. Aerial visual grounding has significant potential applications in areas such as emergency rescue, logistics, and ecological monitoring, which calls for further research and development. Specifically, the ability to manage multiple objects of the same type and their spatial relationships in complex environments remains underdeveloped, leading to limited generalization capabilities in real-world scenarios.

To this end, we propose a more challenging task, termed Aerial Visual Grounding (AerialVG). AerialVG aims to apply the visual grounding task to UAV platforms, and promote the development of UAV intelligent perception and autonomous navigation technologies. Compared with traditional VG tasks, AerialVG faces more challenges, mainly in the following two aspects:
\begin{itemize}[left=0pt]
    \item \textbf{Broad field of view and small objects.} Unlike images from the ground perspective, when UAVs capture images from the air, the field of view is usually much broader, and the size of objects in the image tends to be relatively small. While a larger field of view provides more context and coverage, it also imposes challenges to locate and identify objects at a finer scale.
    \item \textbf{Complex object distribution and spatial relationship.} From the perspective of UAVs, object types are mainly pedestrians and vehicles, which are usually distributed in crowded urban environments. Especially in busy urban streets or traffic-intensive areas, there may be multiple similar or identical objects in the field of view at the same time. For example, multiple similar vehicles are often simultaneously presented in one street. In this case, it is difficult to accurately distinguish the target object based on appearance features alone, and it is necessary to rely on the spatial relationship between objects, such as relative position, distance, and direction, to achieve precise positioning and recognition.
\end{itemize}

To address above issues, we propose the first dataset designed specifically for the AerialVG task. The dataset consists of 5,000 high-resolution images covering a variety of complex urban environments and contains a rich variety of object types, such as pedestrians, vehicles, and other common urban objects. In addition, the dataset includes nearly 50k manual annotations. These annotations not only provide the bounding box of each object but also include natural language descriptions related to the object, which can help the model understand the positional relationship between objects. With these high-quality annotations, a comprehensive and efficient test platform is provided for the AerialVG task.

Aiming at the core challenges in the AerialVG task, we also design an innovative model that can accurately locate objects based on the positional relation between them. Specifically, a Hierarchical Cross-Attention is first proposed to help the model focus more effectively on regions where objects are most likely to appear, thus addressing the issue of vast visual areas. Furthermore, we devise a Relation-Aware Grounding module to understand spatial information such as the relative position, direction, and distance between objects, effectively distinguishing multiple similar objects.

Overall, the main contributions of this work are summarized as follows:
\begin{itemize}[left=0pt]
\setlength{\itemsep}{0pt} 
\item We propose the AerialVG task and introduce the first dataset, providing a valuable resource for advancing research in aerial visual grounding.

\item We present a novel AerialVG model that effectively addresses two key challenges, \emph{i.e.,} the large field of view with small-sized objects and the complex distribution and spatial relationships between objects.

\item We conduct extensive experiments to thoroughly validate the effectiveness of our proposed model, demonstrating its superior performance. 

\end{itemize}

\begin{figure*}[!ht]
    \centering
    \includegraphics[width=1\linewidth]{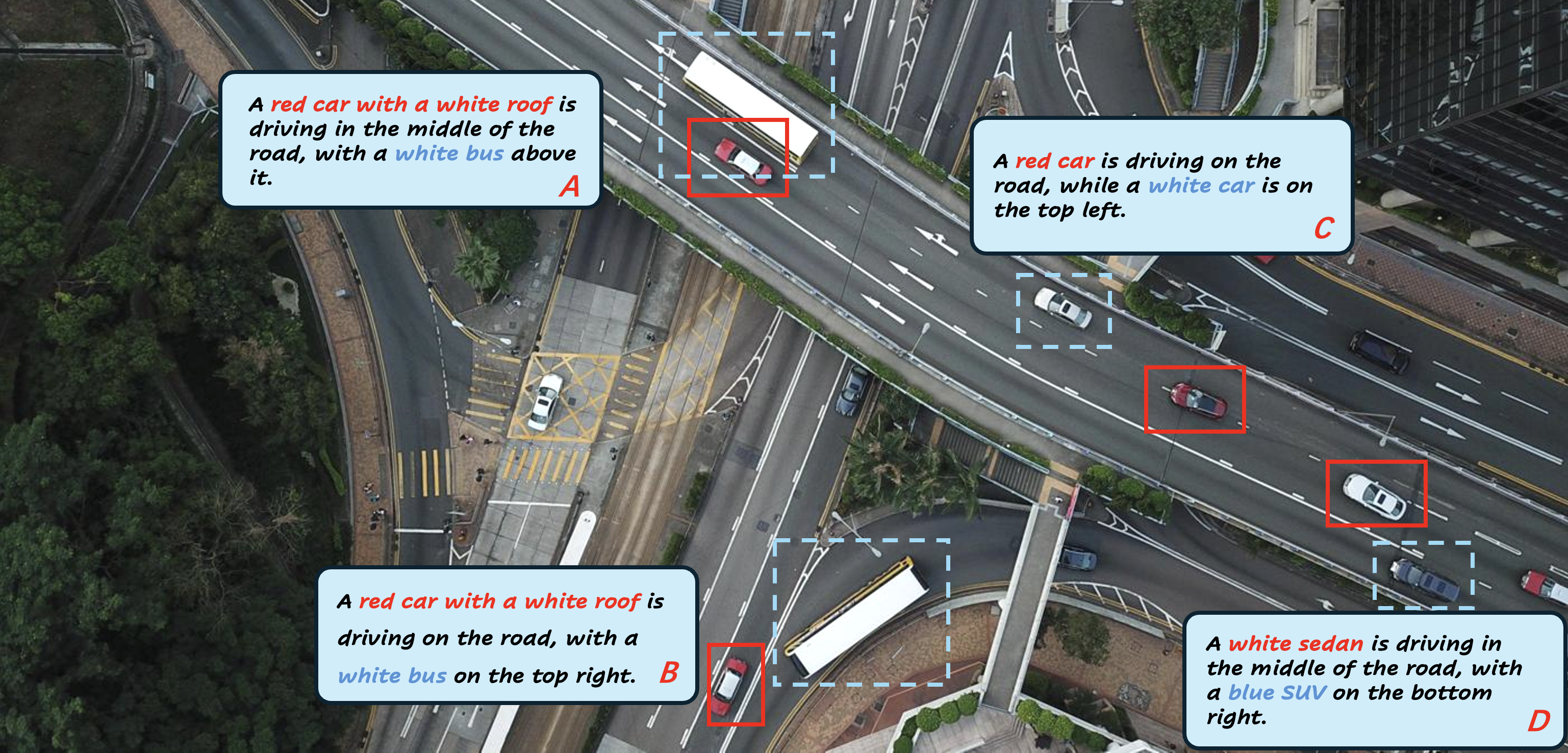}
    \caption{\textbf{Example of AerialVG Dataset.} There are many red cars in the picture. In order to achieve accurate object positioning, we need to use the positional relationship of surrounding auxiliary objects to assist reasoning. To this end, we accurately annotated the data to ensure that the spatial relationship between objects can be fully captured, thereby improving the accuracy of AerialVG.}
    \label{fig:dataset}
\end{figure*}

%% file: sec/2_related_work.tex
\section{Related Work}

\textbf{Visual Grounding.} Visual Grounding (VG) is a fundamental task that requires identifying specific objects in an image based on natural language descriptions. This task has made some progress in both 2D \cite{groundingdino,dai2024simvg} and 3D \cite{tangyiwen,3dvg, guo2023point,tang2024point,tang2024any2point,zhang2023monodetr,tang2022detection} fields. Existing visual grounding methods can be broadly divided into two categories, \emph{i.e.,} (1) two-stage methods\cite{fukui2016multimodal,wang2018learning,liu2020learning} and (2) one-stage methods\cite{ren2016faster,liu2016ssd,redmon2016you}. The former first generates region-level candidate boxes and then uses language to select the best matching region. The latter directly interacts with visual and language cues through bounding boxes, avoiding the computationally intensive proposal box generation and region feature extraction steps present in two-stage methods. Several recent methods\cite{VisualBERT,groundingdino,transvg,ding2021visionlanguagetransformerquerygeneration} have adopted the transformer structure for multimodal fusion, thus achieving better performance. In addition, with the booming development of vision language models (VLMs), some studies \cite{florence2,mplug2,x2vlm} have improved the performance of vision grounding by fine-tuning pre-trained VLMs, making full use of the knowledge and advantages of pre-trained models. These advances have improved both the accuracy and efficiency of VG systems, enabling more complex applications. However, they still cannot solve the spatial relationship problem in AerialVG.

\textbf{Vision-Language Models.} Vision-Language Models (VLMs) integrate visual and linguistic information to perform tasks that require both image understanding and natural language processing. Typically, they use pretrained visual backbones to encode images, and large-scale language models to interpret textual inputs. The visual and textual information is then fused through mechanisms like cross-attention layers. Recent advancements, such as GPT-4o \cite{gpt4o}, LLaVA \cite{llava}, and InstructBLIP \cite{instructionblip}, have achieved significant progress in language-guided navigation and visual reasoning. VLMs have also been adapted to a variety of domains. Specifically, robot action generation produces robot actions based on textual instructions \cite{spatialvla, tracevla}, vision language navigation guides UAVs in aerial environments \cite{openfly, aerialvln}, 3D generation creates and understands 3D data based on language inputs \cite{eeal, pointllm}. Moreover, VLMs are increasingly used in video understanding, biomedical imaging, and remote sensing, demonstrating their versatility across different tasks. These developments highlight the growing potential of VLMs in various real-world applications.

\textbf{Spatial Reasoning Representation.} Spatial reasoning, traditionally part of broader tasks like SLAM \cite{durrant2006simultaneous, cadena2016past} or depth estimation \cite{fu2018deep}, has seen increasing interest in the context of VLMs. Earlier research focused on explicit spatial scene representations, such as spatial memory \cite{gordon2018iqa, gervet2023navigating} or spatial scene graphs \cite{wald2020ssg, hildebr2020scene}, to help machines understand and reason about spatial relations. Recent work has further advanced the spatial reasoning capabilities of VLMs. Some approaches leverage 3D information to enhance VLM performance. \cite{hong20233dconceptlearningreasoning, hong20233d, qu2024livescene, chen2024freegaussian} focus on reconstructing scenes from multi-view images and integrating dense semantic features to improve these representations. These 3D representations are then fused into LLMs for better spatial understanding. However, multi-view images are not always available, and the explicit construction of scenes using dense semantic features can be computationally intensive. Furthermore, the gap between 3D representations and language can often degrade performance. In contrast, other approaches utilize 2D VLMs to understand spatial relationships and measure distances. Unlike the 3D methods, spatial understanding in these models is implicitly encoded, allowing the VLM to directly address spatial reasoning tasks without requiring explicit 3D representations or scene graphs, as seen in works like \cite{spatialvlm, spatialrgpt}.

%% file: sec/3_dataset.tex
\section{Dataset}

\begin{figure}[t]
    \centering
    \includegraphics[width=1\linewidth]{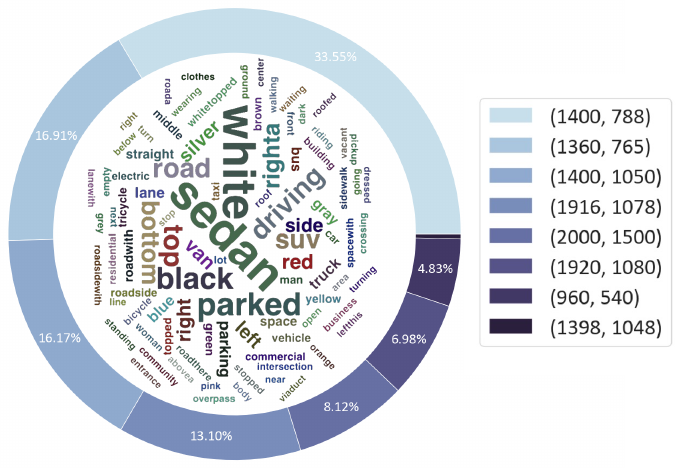}
    \vspace{-2ex}
    \caption{\textbf{Resolution distribution and word cloud of AerialVG dataset.} Most of the images in AerialVG dataset are high-resolution, and the most frequently appearing words are vehicle type, color, and location.}
    \label{fig:size}
    \vspace{-2ex}
\end{figure}

In this paper, we develop the AerialVG dataset as shown in Figure \ref{fig:dataset}. This dataset contains 5,000 high-definition images taken from UAVs.  We annotated them to ensure that they are suitable for aerial visual grounding task. In addition, we provide a specially designed natural language description for each object, which details the objects in the image and their relative positions. The AerialVG dataset provides a solid foundation for fine-grained object grounding in complex scenes from a high-altitude perspective, and especially lays a new research direction for the efficient grounding and understanding tasks of intelligent UAVs.

\subsection{Dataset generation}

The images in our AerialVG dataset are sourced from the VisDrone2019 dataset~\cite{visdrone}. VisDrone consists of 288 video clips (261,908 frames) and 10,209 static images captured by various UAV-mounted cameras across 14 cities, covering a range of environments such as urban and rural areas. This diversity in scene type, location, and conditions (including different weather and lighting) offers a robust foundation for visual grounding tasks. The dataset includes more than 2.6 million manually annotated bounding boxes, highlighting common objects like pedestrians, vehicles, and bicycles, with attributes such as visibility, object class, and occlusion.

In UAV-view images, buildings and roads occupy most of the visual space, while the primary objects of interest are pedestrians and vehicles. They often exhibit similar appearances. Relying solely on coarse-grained categories is insufficient for accurate object localization. Therefore, the annotations must capture fine-grained object features, \emph{e.g.,} color, vehicle type, clothing, and consider the spatial relationships between target objects and prominent surrounding elements. Due to the complexity of visual annotation, we opted for manual annotation to ensure precision.

Our manual annotation process follows these principles:
\begin{itemize}[left=0pt]
\setlength{\itemsep}{0pt} 
\item \textbf{Observer-perspective spatial relationships.} For the convenience of annotation and unification, our position relation is described from the first-person perspective and is divided into eight categories, \emph{i.e.,} above, below, left, right, top left, top right, bottom left, and bottom right. As shown in Figure \ref{fig:dataset} A, the white bus is above the red car with a white roof, not on the right. 

\item \textbf{Selection of prominent auxiliary objects.} When locating the target object, we give priority to selecting the closest object as the auxiliary object to improve positioning accuracy. To ensure simplicity, if one auxiliary object is sufficient to locate the target, no redundant description will be provided. For instance, as shown in Figure \ref{fig:dataset} D, the target object is the white sedan, and we select the blue SUV as the auxiliary object, while disregarding the red car at the top left.
\end{itemize}

\subsection{Dataset statistics}

This subsection presents key statistics of the AerialVG dataset and offers a comparative comparison with existing visual grounding datasets.

\begin{figure}[t]
    \centering
    \includegraphics[width=0.8\linewidth]{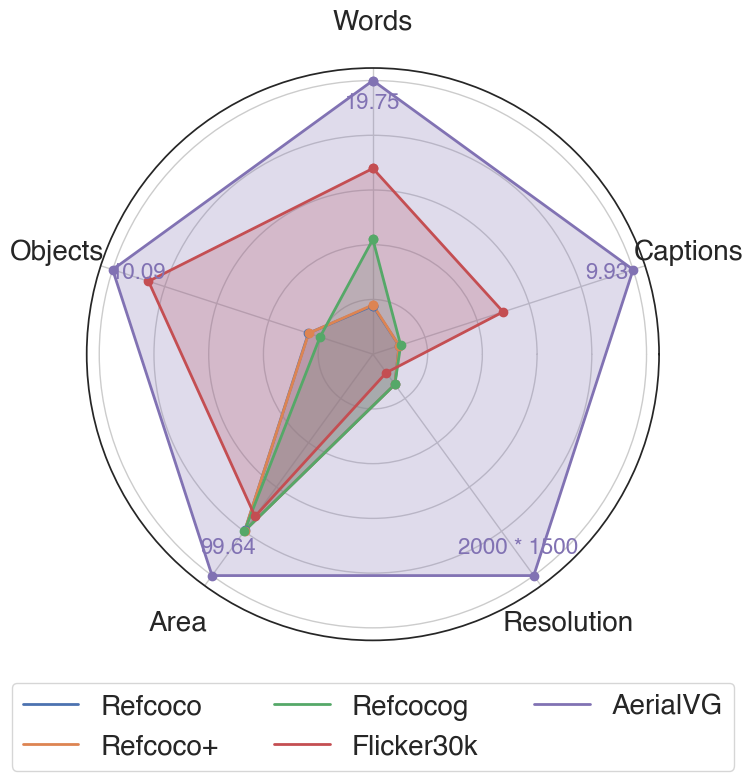}
    \caption{\textbf{Comparison between different datasets.} Words refers to the average number of words in the text; Objects refer to the average number of objects in each image;  Area refers to the proportion of irrelevant areas; Resolution refers to the maximum resolution of the dataset images and Captions refers to the average number of annotations contained in each image.}
    \label{fig:data}
    \vspace{-2ex}
\end{figure}

Figure \ref{fig:size}  shows the resolution distribution and word cloud for the AerialVG dataset. Unlike the RefCOCO series, where over 70\% of images have a resolution around $600\times 650$, the AerialVG dataset consists mostly of images with resolutions greater than $1000\times 500$. This higher resolution and wider field of view make the Aerial VG task more challenging.

The word cloud analysis highlights frequent mentions of vehicle types, colors, and spatial positions. These annotations emphasize the importance of understanding not only object identification but also their relative locations in the environment. This rich annotation style is critical for tasks like UAV navigation, emergency response, and large-scale monitoring, where spatial relationships are essential for accurate decision-making.

Figure \ref{fig:data} compares the AerialVG dataset with existing visual grounding datasets across three key aspects, \emph{i.e.,} the average number of words in the text, the average number of objects located in the image, and the proportion of irrelevant regions. In all three aspects, the AerialVG dataset demonstrates a higher level of challenge, highlighting its complexity and suitability for more advanced visual grounding tasks.

\begin{figure*}[ht]
    \centering
    \includegraphics[width=0.95\linewidth]{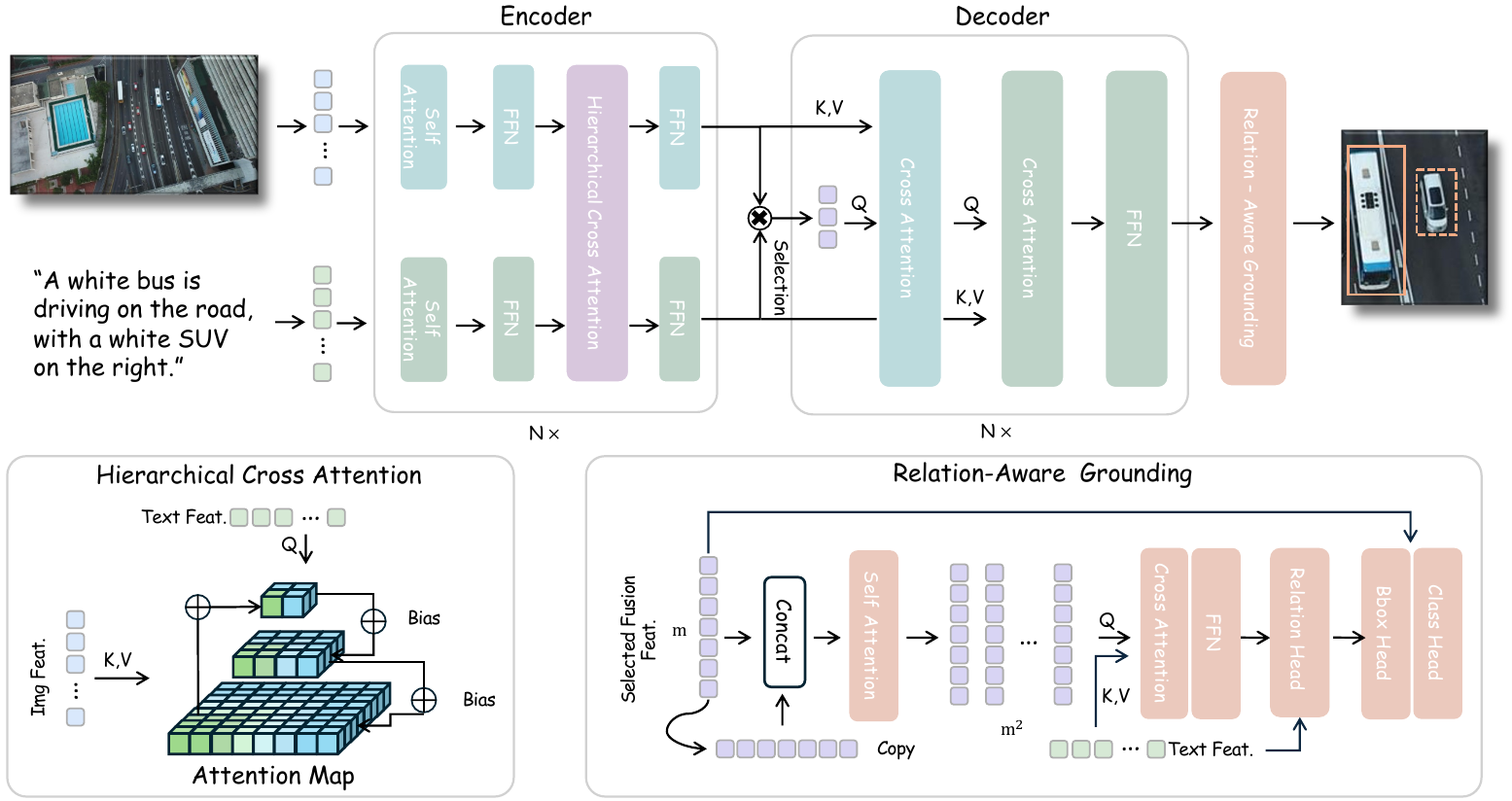}
    \vspace{-1.5ex}
    \caption{\textbf{The architecture of AerialVG model.} Hierarchical Cross Attention directs the model’s focus to potential object locations, while the Relation-Aware Grounding module enables the model to perceive spatial relationships between objects.}
    \label{fig:method}
    \vspace{-2ex}
\end{figure*}

%% file: sec/4_method.tex
\section{Method}

As shown in Figure \ref{fig:method}, we take GroundingDINO~\cite{groundingdino} as the baseline and design an end-to-end model for AerialVG. GroundingDINO~\cite{groundingdino} is built on a transformer architecture, where the encoder stacks self-attention and cross-attention layers to fuse image and text features. After fusing these multimodal features, it selects the most relevant features as queries, which are highly similar to the text. These queries are then decoded in the decoder, and the final bounding boxes and logits are generated by the bbox head and class head, respectively.

Unlike GroundingDINO~\cite{groundingdino}, after embedding the image into hierarchical features using Swin-Transformer~\cite{swintransformer}, we fuse these features with text features encoded by BERT~\cite{bert} via Hierarchical Cross Attention, enabling the model to focus more on potential regions where objects are likely to appear. After the Decoder, we introduce the Relation-Aware Grounding module to capture the positional relationships between different objects.

\subsection{Hierarchical  Cross-Attention}

In aerial view images, high resolution often comes with large portions of the scene occupied by buildings, roads, and other background elements, reducing the focus on key objects. To address this, we propose Hierarchical Cross-Attention, a mechanism designed to help the model more efficiently attend to relevant areas within the scene.

With the hierarchical feature representation enabled by Swin-Transformer\cite{swintransformer}, image features are organized in a layered manner. However, traditional cross-attention methods typically compute a complete attention map by flattening the features for processing, which can lead to information loss during the subsequent fusion process. In hierarchical structures, features at the same spatial location across different layers still represent the same object. Therefore, if a high-attention region is identified in one layer based on textual cues, this region should receive increased attention in subsequent layers. Conversely, if a region shows low attention, it can be partially disregarded in the next layer, enabling the model to prioritize more relevant areas effectively.

As shown in Figure \ref{fig:method}, after calculating the attention map between textual and image features, we retain their hierarchical structure. As the layer number increases, the attention map will also become smaller, making small objects receive insufficient attention in the higher layers. To alleviate this problem, we first convolve the attention map of the first layer to make it consistent with the shape of the last layer map, and fuse the result with it.

\begin{equation}
    A^{*}_{n} = (1-\beta) \times A_{n} + \beta \times C(A_{1}),
    \label{eq:cross2}
\end{equation}
where $C(\cdot)$ denotes the convolution operation, $\beta$ is a hyper-parameter, $A_{j}$ represents the attention map of the j-th layer. The processed attention map is then combined with the highest layer's attention map $A_{n}$, ensuring that even small objects are given adequate focus. After that, we use Equation \ref{eq:cross} to enhance the attention map in lower layers based on the higher layers. 
\begin{equation}
    A^{*}_{i} = (1-\alpha) \times A_{i} + \alpha \times F(A_{i+1}). \label{eq:cross}
\end{equation}
Specifically, we use bilinear interpolation $F(\cdot)$, to expand the attention map $A_{i+1}$ of the $i+1$ layer to the shape of the $i$ layer $A_{i}$. The expanded attention map is then weighted by the hyper-parameter $\alpha$ and combined with $A_{i}$ to form the updated $A^{*}_{i}$. This approach effectively achieves our goal without introducing additional parameters, yielding improvements in performance.

\subsection{Relation-Aware Grounding}

Following GroundingDINO~\cite{groundingdino}, we first perform language-guided query selection after the encoder, and then pass the selected queries into the decoder.  In the decoder, we perform cross-attention on the query separately with the image features and text features to enhance the model’s ability to localize objects. However, experiments showed that these features did not adequately capture the positional relations described in the text.

We draw inspiration from certain 3D visual grounding methods for determining spatial relationships. However, those methods generally require explicit bounding boxes and clearly defined spatial relationships, rather than leveraging full textual descriptions. Our approach circumvents these requirements while still achieving competitive results.

Specifically, we design a Relation-Aware Grounding module to capture positional relation information. As shown in figure \ref{fig:method}, after passing through the decoder, the fused features already contain the positional information of each object. Assuming we have $m$  features before the Relation-Aware Grounding module, we then concatenate every two of the $m$ features in pairs to obtain $m^2$  features, which effectively captures the pairwise relationships between the objects and enhances the model’s ability to learn these interactions. The self-attention operation is performed on $m$ concatenated features in the same column to perceive the positional relation between objects. Subsequently, we conduct cross-attention with $m^2$ queries and the text features. This step enables the model to better align and integrate both visual and textual information, enhancing its understanding of the contextual relationships between objects.

Finally, we first map the last state of the hidden layer to the specified feature dimension through the Relation Head. On this basis, we calculate the similarity between these features and the text features. The calculated similarity scores are then subjected to a softmax operation to generate a set of normalized similarity values. These normalized values represent the relationship score between each pair of objects, that is, how well they match the relationship described in the text. In this way, we can get an $m \times m$ relationship matrix, which represents the relationship score between each pair of objects. We take the maximum value from each row of this relationship matrix as the final relation score for each object. We combine the relation scores with the scores obtained from the query through the class head to form the final logits.

\subsection{Loss Function}
\setlength{\abovedisplayskip}{5pt}  
\setlength{\belowdisplayskip}{5pt}  

Our loss function consists of three parts, \emph{i.e.,} $\mathcal{L}_{\text{reg}}$, $\mathcal{L}_{\text{cls}}$ and $\mathcal{L}_{\text{rel}}$ as shown in Equation \ref{eq:loss}. 
\begin{equation}
\mathcal{L} = \mathcal{L}_{\text{reg}} + \mathcal{L}_{\text{cls}} + \mathcal{L}_{\text{rel}}.
\label{eq:loss}
\end{equation}
Following previous work on GroundingDINO, we employ L1 loss and GIOU loss for bounding box regression in $\mathcal{L}_{\text{reg}}$. For classification, we adopt a contrastive loss between predicted objects and language tokens as $\mathcal{L}_{\text{cls}}$, in line with the GLIP~\cite{zhang2022glipv2} approach.

Additionally, we compute a Relation Loss based on the predicted relationships between object pairs. 
\begin{equation}
\mathcal{L}_{\text{rel}} =  - \log \left( \frac{e^{m_{i,j}}}{\sum_{i,j} e^{m_{i,j}}} \right), \label{eq:relation}
\end{equation}
where Hungarian matching is performed to identify the corresponding target object $m_i$ and the auxiliary object $m_j$. In this way, the element $m_{i,j}$ of the $m \times m$ relationship matrix represents the ground truth relation between the objects. We then use Equation \ref{eq:relation} to maximize the score at the corresponding position $m_{i,j}$ in the relation matrix.

%% file: sec/5_experiment.tex
\section{Experiment}

\subsection{Implementation Details}

Our proposed  AerialVG model, utilizes Swin-L~\cite{swintransformer} as the image backbone. For text representation, we employ the BERT-base~\cite{bert} from Hugging Face as the text backbone. Following Grounding DINO, we extract four levels of image feature scales from the image backbone, ranging from 4× to 32×. We set the maximum number of text tokens to 256. We set $N=6$ for Encoder and Decoder layers,  while the Relation-Aware Grounding module consists of three layers. The final loss function consists of multiple components, \emph{i.e.,} classification loss (or contrastive loss), L1 loss, GIoU loss, and relation loss, with corresponding weights set to 1.0, 5.0, 2.0, and 1.0, respectively.  The hyper-parameters $\alpha$ and $\beta$ are set to 0.2 and 0.3. We conduct extensive experiments on the AerialVG dataset and the RefCOCO, RefCOCO+, and RefCOCOg datasets~\cite{refcoco}. Additionally, we perform ablation studies to demonstrate the effectiveness of our model design.

Our model training consists of two stages. Initially, we train the model without the Relation-aware Grounding module, using existing VG datasets following Grounding DINO \cite{groundingdino} to achieve convergence. Once the model has converged, we freeze all other parameters and train the Relation-Aware Grounding module using our AerialVG dataset only, allowing it to specifically capture inter-object relationships. More details can be found in our supplementary material.

\subsection{Evaluation metric}

In AerialVG, object homogeneity is a significant challenge, as many objects share nearly identical visual characteristics. A model that performs well on Top-5 Accuracy but poorly on Top-1 Accuracy likely relies primarily on visual similarity rather than understanding the spatial relationships or contextual cues that distinguish between objects. By analyzing both Top-1 and Top-5 Accuracy, we can diagnose specific failure modes of the model. Model with both high Top-1 and Top-5 Accuracy demonstrates strong object discrimination and contextual understanding ability, making it more reliable for AerialVG-related applications.

\begin{table}[htbp]
  \centering
  \caption{Performance comparison on the AerialVG dataset.}
    \begin{tabular}{cccccc}
    \toprule
    \multicolumn{2}{c}{\multirow{3}[4]{*}{\textbf{Method}}} & \multicolumn{4}{c}{\textbf{Setting}} \\
\cmidrule{3-6}    \multicolumn{2}{c}{} & \multicolumn{2}{c}{\textbf{Zero-Shot}} & \multicolumn{2}{c}{\textbf{Fine-Tuning}} \\
    \multicolumn{2}{c}{} & \textbf{Top 1} & \textbf{Top  5} & \textbf{Top 1} & \textbf{Top  5} \\
    \midrule
    \multicolumn{2}{c}{TranVG\cite{transvg}} &   2.57    &    3.65   &    11.53   & 13.68 \\
    \multicolumn{2}{c}{Dynamic-MDETR\cite{dai2021dynamic}} &     2.61  &     4.63  &  19.87     & 29.87 \\
    \multicolumn{2}{c}{SimVG\cite{dai2024simvg}} &     3.61  &    4.68   & 20.94  & 31.32 \\
    \multicolumn{2}{c}{Grounding DINO\cite{groundingdino}} & 11.10  & 13.79  & 29.36  & 78.87  \\
    \multicolumn{2}{c}{AerialVG(Ours)} & -     & -     & \textbf{50.01} & \textbf{87.00} \\
    \bottomrule
    \end{tabular}
  \label{tab:aerialvg}%
\end{table}%

\begin{table*}[t!]
  \centering
  \caption{Performance comparison on RefCOCO, RefCOCO+, and RefCOCOg datasets.}
  \label{tab:refcoco}
  \begin{adjustbox}{center}
    \resizebox{0.8\textwidth}{!}{ 
    \renewcommand{\arraystretch}{1.3}
    \begin{tabular}{lcccccccccccc}
    \toprule
    \multirow{2}{*}{\textbf{Method}} & \multirow{2}{*}{\textbf{Backbone}} & \multicolumn{3}{c}{\textbf{RefCOCO}} & \multicolumn{3}{c}{\textbf{RefCOCO+}} & \multicolumn{2}{c}{\textbf{RefCOCOg}} \\
    \cmidrule{3-10}
     &  & \textbf{val} & \textbf{testA} & \textbf{testB} & \textbf{val} & \textbf{testA} & \textbf{testB} & \textbf{val} & \textbf{test} \\
    \midrule
    TranVG \cite{transvg} & ResNet-101 & 80.32 & 82.67 & 78.12 & 63.5 & 68.18 & 55.63 & 67.11 & 67.44 \\
    Dynamic-MDETR \cite{dai2021dynamic}& ResNet-50 & 85.97 & 88.82 & 80.12 & 74.83 & 81.7 & 63.44 & 73.18 & 74.49 \\
    SimVG \cite{dai2024simvg}& ViT-L & \textbf{90.61} & 92.53 & 87.68 & \textbf{85.36} & \textbf{89.61} & \textbf{79.74} & 82.67 & 86.8 \\
    Grounding DINO\cite{groundingdino} & Swin-L & 90.56 & \underline{93.19} & \underline{88.24} & 82.75 & 88.95 & 75.92 & \underline{86.13} & \underline{87.02} \\
    AerialVG (Ours) & Swin-L &  \underline{90.58}& \textbf{93.23} & \textbf{88.56} & \underline{82.83} & \underline{89.02} & \underline{76.03} & \textbf{86.52} & \textbf{88.04} \\
    \bottomrule
    \end{tabular}
    }
  \end{adjustbox}
\end{table*}

\subsection{Results}

Table \ref{tab:aerialvg} presents the evaluation results of different models on the AerialVG dataset. In the Zero-Shot setting, existing VG models exhibit limited performance, with both Top-1 Accuracy and Top-5 Accuracy hovering around 10\%. This result highlights the challenges posed by high-resolution aerial imagery and the unique characteristics of the aerial viewpoint, which are not well-represented in existing datasets. The poor performance of these models further underscores the necessity of AerialVG as a benchmark for evaluating visual grounding in aerial scenes.

After fine-tuning, Top-5 Accuracy shows a significant improvement, indicating that models can better recognize and retrieve relevant object candidates when adapted to the AerialVG. However, Top-1 Accuracy remains relatively low, suggesting that while models are able to identify visually similar objects within the scene, they struggle to distinguish the correct target based on spatial relationships and contextual cues. 



In comparison, the proposed AerialVG model demonstrates superior performance over previous approaches. It demonstrates a better understanding of object relationships in aerial images, resulting in higher Top-1 and Top-5 accuracy. This suggests that our model effectively integrates the Hierarchical Cross-Attention mechanisms and the Relation-Aware Grounding module, which are crucial for distinguishing visually similar objects based on spatial dependencies. These results validate our architectural choices and demonstrate the effectiveness of our model in addressing the unique challenges of aerial visual grounding.

In addition to the AerialVG dataset, we also evaluate our AerialVG model on the RefCOCO\cite{refcoco}, RefCOCO+\cite{refcoco}, and RefCOCOg\cite{refcocog} datasets to assess its performance on standard visual grounding benchmarks. As shown in Table \ref{tab:refcoco}, our AerialVG model also presents good performance across multiple metrics on the RefCOCO series datasets, achieving comparable results with previous state-of-the-art approaches.

These findings confirm that the integration of the Hierarchical Cross-Attention and Relation-Aware Grounding module does not compromise the model’s ability to accurately localize individual objects. Despite being designed to enhance spatial reasoning and relational understanding, these components maintain strong object-level grounding capabilities. This highlights the versatility of our model, making it effective in both aerial imagery and conventional ground-based VG tasks.

\begin{table}[htbp]
  \centering
  \caption{\textbf{Ablation study.} Attn refers to hierarchical cross attention, Relation refers to relation aware grounding.}
    \begin{tabular}{ccrr}
    \toprule
    \multicolumn{2}{c}{\multirow{3}[4]{*}{\textbf{Method}}} & \multicolumn{2}{c}{\textbf{AerialVG}} \\
\cmidrule{3-4}    \multicolumn{2}{c}{} & \multicolumn{2}{c}{\textbf{Fine-Tuning}} \\
    \multicolumn{2}{c}{} & \multicolumn{1}{c}{\textbf{Top 1}} & \multicolumn{1}{c}{\textbf{Top  5}} \\
    \midrule
    \multicolumn{2}{c}{Grounding DINO (Baseline)} &    29.36   & 78.87 \\
    \multicolumn{2}{c}{Grounding DINO + Attn} &    29.78   &  82.32\\
    \multicolumn{2}{c}{Grounding DINO + Relation} &     45.63  & 79.63 \\
    \multicolumn{2}{c}{Grounding DINO + Attn +Relation} &   50.01    & 87.00 \\
    \bottomrule
    \end{tabular}%
  \label{tab:ablation}%
  \vspace{-2ex}
\end{table}%

\subsection{Ablation study}

To further validate the effectiveness of the Hierarchical Cross-Attention  and the Relation-Aware Grounding module, we conduct ablation experiments, as shown in Table  \ref{tab:ablation}. The results demonstrate that each component plays a crucial role in improving different aspects of the model’s performance.
\begin{itemize}[left=0pt]
\setlength{\itemsep}{0pt} 
    \item Hierarchical Cross-Attention significantly enhances Top-5 Accuracy, indicating that this module helps the model focus on fine-grained details within the image. By attending to multi-level image features, the model can better capture subtle differences between similar objects, leading to improved candidate retrieval.
    \item Relation-Aware Grounding module notably improves Top-1 Accuracy, demonstrating its ability to refine object selection by incorporating relational reasoning. This confirms that the module helps disambiguate objects based on spatial and contextual relationships, rather than relying solely on visual appearance.
\end{itemize}

\begin{figure}[h]
    \centering
    \includegraphics[width=\linewidth]{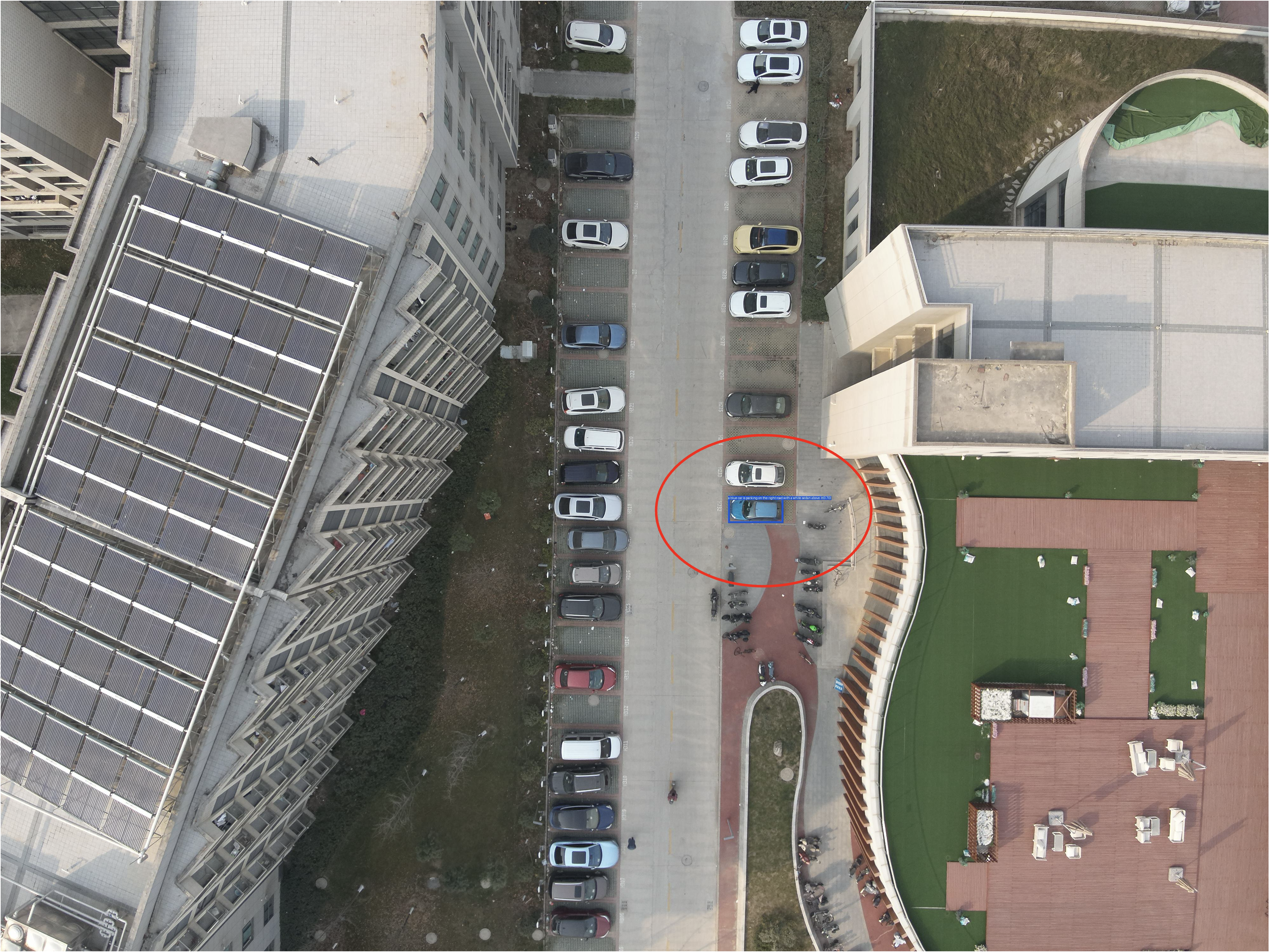}
    \caption{\textbf{Example of Qualitative Test.} Annotation: A blue car is parking on the right road with a white sedan above it. The model output score for this result is 0.70.}
    \label{fig:demo}
    \vspace{-2ex}
\end{figure}

\subsection{Qualitative Results}


To better validate the generalization ability of our model, we conducted real-world experiments by capturing a set of aerial images using the DJI M30T drone. The images were taken at an altitude of 50 meters with a resolution of 4000×3000. The images are filled with various types of vehicles, posing a significant challenge to object localization. However, as shown in Figure \ref{fig:demo}, our model still successfully locates the objects accurately based on spatial relationship information, demonstrating its strong generalization capability. We will show more visualization results in the supplementary material.

%% file: sec/6_conclusion.tex
\section{Limitations and Future Work}

The primary limitation of current visual grounding models, including our AerialVG, is their inability to effectively recognize unstructured background information, such as roads and other environmental elements. While these models excel at detecting foreground objects, they often struggle to incorporate contextual information from the background, which is crucial for tasks requiring a more holistic understanding of the scene. Promoting the spatial intelligence of perception models will be a key focus of our future work.

\section{Conclusion}

In this paper, we introduce AerialVG, a visual grounding task designed specifically for aerial perspectives. We construct the first real-world dataset tailored for this task, featuring high-resolution aerial imagery with densely populated scenes and complex spatial relationships. To address the challenges of fine-grained object differentiation and spatial reasoning, we propose an effective model, integrating Hierarchical Cross-Attention and a Relation-Aware Grounding module to enhance object localization accuracy. Extensive experiments validate the effectiveness of our approach, demonstrating state-of-the-art performance on the AerialVG dataset and competitive results on standard RefCOCO benchmarks. Our findings highlight the importance of spatial reasoning in aerial visual grounding and establish AerialVG as a challenging benchmark for future research. We hope that our dataset and the proposed method will serve as a solid foundation for advancing aerial view perception and autonomous vision-and-language applications.